# Enhancing Keyphrase Generation by BART Finetuning with Splitting and Shuffling

Bin CHEN and Mizuho IWAIHARA[*]

Graduate School of Information, Production, and Systems, Waseda University
Kitakyushu 808-0135, Japan
`chenbin@asagi.waseda.jp iwaihara@waseda.jp`

**Abstract.** Keyphrase generation is a task of identifying a set of phrases that best represent the main topics or themes of a given text. Keyphrases are dividend int present and absent keyphrases. Recent approaches utilizing sequence-to-sequence models show effectiveness on absent keyphrase generation. However, the performance is still limited due to the hardness of finding absent keyphrases. In this paper, we propose Keyphrase-Focused BART, which exploits the differences between present and absent keyphrase generations, and performs finetuning of two separate BART models for present and absent keyphrases. We further show effective approaches of shuffling keyphrases and candidate keyphrase ranking. For absent keyphrases, our Keyphrase-Focused BART achieved new state-of-the-art score on F1@5 in two out of five keyphrase generation benchmark datasets.

**Keywords:** keyphrase generation, deep learning, BART Finetuning, generative language model

## 1 Introduction

Keyphrase generation is an important task that involves identifying a set of terms or phrases that best represent the main topics or themes of a given text, having applications in information retrieval, document classification, and summarization. A **present keyphrase** is such that its word sequence appears in the document with its order preserved. Present keyphrases can be extracted from the document. An **absent keyphrase** is not present in the text but relevant to the topic of the document.

Keyphrase extraction has been extensively studied [1,6,9]. However, extractive methods cannot find absent keyphrases that have not appeared in the article. Recent generative methods, such as CopyRNN [5] and CatSeq [13], can directly generate candidate present and absent keyphrases from input document representations.

BART is a pre-trained generative language model based on a denoising autoencoder [7], which can directly perform sequence generation tasks through finetuning, which can be applied to keyphrase generation [6].

---

[*] corresponding author

We point out that in most of the previous work based on generative language models, finetuning is done on present and absent keyphrases together [6]. However, we argue that there exist considerable differences in the tasks of extracting present keyphrases and generating absent keyphrases, which motivates us to propose splitting the absent and present keyphrase generation tasks into two parts, and train two different generative models, where different hyperparameters are used for finetuning.

The main contributions of this paper are: (1) A new model *Keyphrase-Focused BART* is proposed, in which two BART models are finetuned separately on present and absent keyphrases, with different hyperparameter settings. (2) Shuffling keyphrase lists for prompting order-independence and augmenting samples is proposed. (3) A keyphrase ranker by a BERT cross-encoder combined with TF-IDF is introduced to improve keyphrases generated by the BART models. (4) Our experimental evaluation confirms effectiveness of these approaches. Our proposed Keyphrase-Focused BART shows new state-of-the-art records on absent keyphrases, on datasets SemEval and KP20K on F1@5. The ratio of F1@5 over the previous state-of-the-art is ranging between 9 to 37 percent, showing a wide improvement.

## 2 Related Work

The following models are representative generative models, and compared against our proposed model in our evaluations:

**CatSeq** [13]: An RNN-based sequence-to-sequence model with copy mechanism trained under ONE2SEQ paradigm.

**CatSeqTG-2RF1** [2]: Based on CatSeq with title encoding and cross-attention.

**GAN$_{MR}$** [10]: RL-based fine-tuning extension on CatSeq.

**Fast and Constrained Absent KG** [11]: Prompt-based keyphrase generation methods, with prompt created around keyword and apply mask predict decoder.

**ONE2SET** [12]: A sequence-to-sequence model based on transformers. ONE2SET generates a set of keyphrases, where the keyphrase order is ignored.

**ONE2SET+KPDrop-a** [4]: KRDrop randomly drops present keyphrases for enhancing absent keyphrase generation.

**ChatGPT** [8]: The large language model ChatGPT is instructed to generate keyphrases.

## 3 Keyphrase-Focused BART

Fig. 1 shows our proposed model **Keyphrase-Focused BART**, which has two generative pretrained language models finetuned separately on present and absent keyphrases.

**Language model separation**: In the existing approaches [2,4,10,11,12,13] of keyphrase generation by generative language models, just a single language model is trained over the union of present and absent keyphrases. KPDrop [5] randomly masks present keyphrases to be used as augmentation for absent keyphrases, where the absent prediction could be enhanced when the absent and masked phrases are semanti-cally similar. But keyphrases are often topically distinct each other. Also, absent phrases need to be chosen from candidates that are vastly larger than the present phrases. The imbalanced candidate spaces for present and absent keyphrases will cause differences in the optimum training processes for both types.

To resolve the above issues, we introduce an architecture in which two separated BART models are trained independently, where one model is trained only by present keyphrases, while the other model is trained only by absent keyphrases. Different hyperparameter settings are used for these BART models, to separately optimize the learning processes for the two tasks.

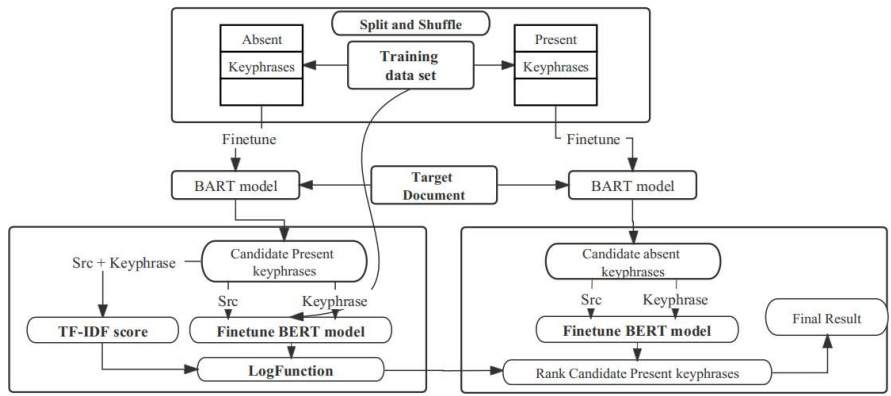

**Fig. 1.** Model Architecture of Keyphrase-Focused BART

**Shuffling and expanding**: Keyphrase lists shall be order independent. In [12], it is mentioned that the BART model might try to generate keyphrases by considering contextual relationships between the keyphrases. To reduce contextualities in learning output sequences, we apply shuffling on the training keyphrase lists, and add the shuffled sequences to the training dataset.

**Ranking by BERT cross-encoder**: We formulate ranking candidate keyphrases as a binary classification task such that the reference keyphrases are labeled as 1, otherwise 0. The confidence score of a finetuned BERT cross-encoder [3] is is coupled with TF-IDF score as: $log\, Score = [\alpha * \log Cross + (1-\alpha) * \log tf\_idf]$, where parameter $\alpha$ is set to 0.7 in this paper. Note that TF-IDF score is not applicable for absent keyphrases.

## 4 Experiments

### 4.1 Experimental settings

We perform experiments on the five widely-used benchmark keyphrase datasets [12]: Inspec, Krapivin, NUS, SemEval, and KP20K. The baseline models we compared are those listed in bold fonts in Section 2. Below lists variations of our model, evaluated as ablations in the experiments:

**Basic BART**: BART model finetuned on the union of present and absent key-phrases.

**A-P Separate**: Two BART models are finetuned on 1) present keyphrases only, with 4 training epochs, and 2) absent keyphrases only, and 8 training epochs, where more epochs are allocated than present keyphrase model, to deal with slow convergence. The learning rate and batch size are 1e-5 and 12, respectively, for both BART models. No shuffling on keyphrase lists is done.

**A-P Separate+Shuffle(1)**: A-P Separate, and shuffling phrase lists once and add new lists into training dataset. The final dataset KP20K increased from 514,154 to 848,684.

**A-P Separate+Shuffle(2)**: A-P Separate, and shuffling phrase lists twice and add new lists into training dataset. The final dataset KP20K increased from 514,154 to 1,086,979.

**A-P Separate+Shuffle(1)+Rank**: A-P Separate+Shuffle(1), and then ranking by the BERT cross encoder. Its hyperparameter settings are: learning rate 5e-6, batch size 24, and training epochs 3. Negative filtering is used which is removing correctly predicted negatives after each epoch.

We follow [2,4] on evaluation metrics. For present and absent keyphrases, we use macro-average $F1@5$ and $F1@M$. $F1@M$ takes into account all the keyphrases generated by the model and compares them to the reference keyphrases.

The results are shown in Table 1 and Table 2. All the results of our models are obtained by averaging four runs. The results of the baselines are from the cited papers.

### 4.2 Results and analysis

**Results on Present Keyphrases:** From Table 1, we can see that A-P Separate that separates the training dataset shows improves over Basic BART. By adding shuffling and separating to A-P Separate, the F1 scores of present keyphrases are further improved compared to using the basic BART model directly, but there is still a gap compared to ONE2SET [12]. Then the model A-P Separate + Shuffle(1) + Rank that uses the ranking unit by BERT cross-encoder and TF-IDF is further improving performance, and achieving highest F1@5 result on the Inspec dataset. We find that shuffling twice is rather falling behind of shuffling once, so we choose A-P Separate + Shuffle(1) + Rank as our best model for present keyphrases.

**Results on Absent Keyphrases:** The results on absent keyphrases are shown in Table 2. We find that shuffling keyphrases once and expanding a dataset is showing improvements of 0 – 1.0 percent on F1 score. The ranking unit, on the other hand, shows little or no improvement of -0.3 to +0.1 percent to the model without the ranking unit.

Overall, our Keyphrase-focused BART, with configuration of A-P Separate + Shuffle (1), achieved new state-of-the-art results on SemEval and KP20K on F1@5. The improvement of F1@5 over ONE2SET-KPDrop-a is ranging between 9 to 37 percent, achieving wide improvements. ChatGPT is showing highest score on Inspec, but the scores reported in [8] are falling behind of our proposed model on the other three datasets.

**Table 1.** Results on Present Keyphrases (F1-score× 100 )

| | Inspec | | NUS | | Krapivin | | SemEval | | KP20K | |
|---|---|---|---|---|---|---|---|---|---|---|
| **Model** | F1@5 | F1@M | F1@5 | F1@M | F1@5 | F1@M | F1@5 | F1@M | F1@5 | F1@M |
| CatSeq [13] | 22.5 | 26.2 | 32.3 | 39.7 | 26.9 | 35.4 | 24.2 | 28.3 | 29.1 | 36.7 |
| CatSeqTG-2RF1 [2] | 25.3 | 30.1 | 37.5 | 43.3 | 30.0 | 36.9 | 28.7 | 32.9 | 32.1 | 38.6 |
| GANMR [10] | 25.8 | 29.9 | 34.8 | 41.7 | 28.8 | 36.9 | - | - | 30.3 | 37.8 |
| Fast and Constrained [11] | 26.0 | 29.4 | 41.2 | 43.9 | - | - | 32.9 | 35.6 | 35.1 | 35.5 |
| SET-TRANS (ONE2SET) [12] | 28.5 | 32.4 | 40.6 | **45.1** | 32.6 | **36.4** | 33.1 | **35.7** | 35.9 | 39.2 |
| ONE2SET-KPDrop-a [4] | 29.8 | 30.6 | **42.6** | 44.4 | **34.0** | 35.3 | **33.6** | 34.4 | **38.5** | **39.6** |
| ChatGPT [8] | 32.5 | **40.3** | - | 20.0 | - | - | - | 18.6 | 23.2 | 25.1 |
| Proposed **Keyphrase-Focused BART** finetuned on present keyphrases | | | | | | | | | | |
| Basic BART | 29.5 | 29.5 | 27.1 | 27.1 | 19.9 | 19.9 | 21.4 | 21.4 | 30.7 | 30.7 |
| A-P Separate + NoShuffle | 30.9 | 30.9 | 34.7 | 34.7 | 25.2 | 25.2 | 22.7 | 22.7 | 29.9 | 29.9 |
| A-P Separate + Shuffle(1) | 33.1 | 33.1 | 37.9 | 37.9 | 27.8 | 27.8 | 27.6 | 27.8 | 31.5 | 31.5 |
| A-P Separate + Shuffle(2) | 32.9 | 32.9 | 37.2 | 37.2 | 32.3 | 32.3 | 23.9 | 23.9 | 30.3 | 30.3 |
| **A-P Separate + Shuffle(1) + Rank** | **35.8** | 35.8 | 41.2 | 41.2 | 29.0 | 29.0 | 28.3 | 28.3 | 33.7 | 33.7 |

**Table 2.** Results on Absent Keyphrases (F1-score× 100 )

| | Inspec | | NUS | | Krapivin | | SemEval | | KP20K | |
|---|---|---|---|---|---|---|---|---|---|---|
| **Model** | F1@5 | F1@M | F1@5 | F1@M | F1@5 | F1@M | F1@5 | F1@M | F1@5 | F1@M |
| CatSeq [13] | 0.4 | 0.8 | 1.6 | 2.8 | 1.8 | 3.6 | 1.6 | 2.8 | 1.5 | 3.2 |
| CatSeqTG-2RF1[2] | 1.2 | 2.1 | 1.9 | 3.1 | 3.0 | 5.3 | 2.1 | 3.0 | 2.7 | 5.0 |
| GANMR [10] | 1.3 | 1.9 | 2.6 | 3.8 | 4.2 | 5.7 | - | - | 3.2 | 4.5 |
| Fast and Constrained [11] | 1.7 | 2.2 | 3.6 | 4.2 | - | - | 2.8 | 3.2 | 3.2 | 4.2 |
| SET-TRANS (ONE2SET) [12] | 2.1 | 3.4 | 4.2 | 6.0 | 4.8 | **7.3** | 2.6 | 3.5 | 3.6 | 5.8 |
| ONE2SET-KPDrop-a [4] | 3.2 | 3.2 | **7.4** | **7.4** | **7.2** | 7.2 | 4.6 | 4.7 | 6.5 | 6.6 |
| ChatGPT [8] | **4.9** | **5.9** | - | 4.2 | - | - | - | 2.1 | 4.4 | 5.6 |
| Proposed **Keyphrase-Focused BART** finetuned on absent keyphrases | | | | | | | | | | |
| Basic BART | 2.4 | 2.4 | 3.8 | 3.8 | 3.8 | 3.8 | 2.9 | 2.9 | 6.5 | 6.5 |
| A-P Separate + NoShuffle | 2.4 | 2.4 | 5.6 | 5.6 | 6.1 | 6.1 | 4.5 | 4.5 | 7.9 | 7.9 |
| **A-P Separate + Shuffle(1)** | 2.4 | 2.4 | 5.6 | 5.6 | 6.4 | 6.4 | 4.9 | 4.9 | **8.9** | **8.9** |
| A-P Separate + Shuffle(1) + Rank | 2.3 | 2.3 | 5.6 | 5.6 | 6.0 | 6.0 | **5.0** | **5.0** | 8.8 | 8.8 |

## 5      Conclusion and Future Work

In this paper, we proposed a generative language model approach for keyphrase generation. We show that splitting the generative language model into two tasks of absent keyphrase generation and present keyphrase extraction, and training them separately bring considerable performance improvements. Overall, for absent keyphrase generation, our Keyphrase-focused BART shows improvements on F1@5 by 9 and 37 percent on two datasets, from the previous state-of-the-art model. In future work, we will consider integrating prompt-based approaches for ranking candidate keyphrases.